\documentclass[11pt,a4paper]{article}
\usepackage[hyperref]{acl2021}
\usepackage{times}
\usepackage{graphicx}
\usepackage{latexsym}
\usepackage{hyperref}
\usepackage{multirow}

\usepackage{microtype}

\aclfinalcopy % Uncomment this line for the final submission
\title{IITP at WAT 2021: System description for English-Hindi Multimodal Translation Task}

\author{Baban Gain\thanks{\enspace Equal contribution} \\
  Indian Institute of Technology Patna \\
  Patna, India\\
  \texttt{gainbaban@gmail.com} \\\And
  Dibyanayan Bandyopadhyay\footnotemark[1] \\
  Indian Institute of Technology Patna \\
  Patna, India\\
  \texttt{dibyanayan@gmail.com} \\\AND
    Asif Ekbal \\
  Indian Institute of Technology Patna \\
  Patna, India\\
  \texttt{asif@iitp.ac.in} \\}

\date{}

\begin{document}
\maketitle
\begin{abstract}

\end{abstract}
Neural Machine Translation (NMT) is a predominant machine translation technology nowadays because of its end-to-end trainable flexibility. However, NMT still struggles to translate properly in low-resource settings specifically on distant language pairs. One way to overcome this is to use the information from other modalities if available. The idea is that despite differences in languages, both the source and target language speakers see the same thing and the visual representation of both the source and target is the same, which can positively assist the system. Multimodal information can help the NMT system to improve the translation by removing ambiguity on some phrases or words. We participate in the 8th Workshop on Asian Translation (WAT - 2021) for English-Hindi multimodal translation task and achieve 42.47 and 37.50 BLEU points for Evaluation and Challenge subset, respectively.

\section{Introduction}

Recent progress in neural machine translation (NMT) focuses on translating a source language into a particular target language. Various methods have been proposed for this task and most of them deal with the textual data. There are certain drawbacks while performing machine translation using only textual datasets.

Human performs translation which is based upon language grounding: our sense of meaning emerges from interacting with the world. NMT methods do not have any mechanism to perform language grounding; thus they are devoid of capturing the true meaning of sentences or phrases while translating them into the other languages. For example, it needs to translate the word ``cricket", it can get confused if it is the game cricket or the insect cricket. But the visual information can clear the ambiguity. Multi-modal translation aims to alleviate this issue by training an NMT model on textual data along with associated images to perform language grounding.\\
This shared task deals with developing multi-modal NMT models for English-Hindi translation. The choice of languages depends on the following issues:
\textit{i). } Hindi is the most spoken language in India and the fourth most spoken language in the world with 600 million speakers\footnote{\href{https://www.ethnologue.com/guides/ethnologue200}{https://www.ethnologue.com/guides/ethnologue200}}. Despite the huge amount of speakers, suitable resources in Hindi is limited due to the various factors. \textit{ii) }Automatic translation of texts from one language to the another is a difficult task. Specifically, when one or both of them are resource-poor and distant from each other.\\
In Multimodal NMT (MNMT), information from the other modalities like audio, image, video, etc. are used along with text to generate the translation. In low-resource languages, this is particularly used to improve the low-quality translations as even though vocabularies, grammar of two languages are different but their visual representation is the same. There are several proposed multi-modal methods for translations that exploit the features of the associated image for better translation. State-of-the-art methods might achieve better accuracy than the models we used. Our main motivation for using simplistic models is to demonstrate a proof-of-concept to be used for multi-modal translation among the resource-poor language pairs. We achieved good results on both Challenge and Evaluation set in different evaluation metrics including BLEU, RIBES, AMFM. In subsequent modifications, we aim to develop our models incorporating several state-of-the-art features. The following sections describe our processes in greater details.
 
\begin{table*}[]
\resizebox{\textwidth}{!}{%
\begin{tabular}{lllll} \hline
Dataset      & Type                  & Sentences & Avg length source & Avg length target \\ \hline
Pre-training & Parallel              & 273,885   & 12.33             & 13.36             \\
Train        & parallel + multimodal & 28929     & 4.95              & 5.02              \\
Valid        & parallel + multimodal & 998       & 4.93              & 4.99              \\
Evaluation   & parallel + multimodal & 1595      & 4.92              & 4.92              \\
Challenge    & parallel + multimodal & 1400      & 5.85              & 6.17             
\end{tabular}%
}
\caption{Descriptions of datasets used for our experiments}
\label{tab:datasets}
\end{table*}

\section{Related Works}

There have been many attempts to use information other than the source for better translation. Uni-modal systems include document-level NMT \cite{wang-etal-2017-exploiting-cross}, sentence-level NMT with contextual information \cite{baban-gain-chat-ijcnn}, etc. Among multimodal systems,
\cite{huang-etal-2016-attention} used an object detection system and extracted local and global image features. Thereafter, they used those image features as additional inputs to encoder and decoder. \cite{delbrouck2017modulating} used attention mechanism on visual inputs for the source hidden states. \cite{Lin_2020} used Dynamic Context-guided Capsule Network \cite{sabour2017dynamic} (DCCN) for iterative extraction of related visual features.

Multimodal Machine Translation (MMT) for English-Hindi has not been well explored yet. \cite{dutta-chowdhury-etal-2018-multimodal} used synthetic data for training. Furthermore they used multi-modal, attention-based MMT which incorporate visual features into different parts of both the encoder and the decoder \cite{calixto-liu-2017-incorporating}. \cite{sanayai-meetei-etal-2019-wat2019} used a Recurrent Neural Network (RNN) based approach achieving BLEU score of 28.45 on Evaluation set and 12.58 on Challenge set. \cite{laskar-etal-2020-multimodal} exploited monolingual data for better translation. 
Recent works tried to focus on developing unsupervised model for multi-modal NMT. \citet{DBLP:journals/corr/abs-1811-11365} demonstrated an unsupervised method based on the language translation cycle consistency loss conditional on the image. This is done to learn the bidirectional multi-modal translation simultaneously. Moreover, \citet{SU202147} showed that jointly learning text-image interaction instead of modeling them separately using attentional networks is more useful. This result is in line with several state-of-the-art visual transformer related models, such as VisualBERT \citep{DBLP:journals/corr/abs-1908-03557}, UNITER \citep{DBLP:journals/corr/abs-1909-11740} etc.

\section{Methodology}
\subsection{Dataset Description}
We use Hindi Visual Genome 1.1 dataset \cite{hindi-visual-genome:2019}\cite{nakazawa-etal-2020-overview}\cite{nakazawa-etal-2021-overview} for our experiments. This dataset consists of 28,929 parallel English-Hindi sentence pairs along with the associated images. Furthermore, we use HindEnCorp dataset for pre-training containing 273K English-Hindi sentence pairs without images. Statistics of the datasets are shown in Table \ref{tab:datasets}.

\begin{figure}
\centering
\includegraphics[width=0.5\textwidth]{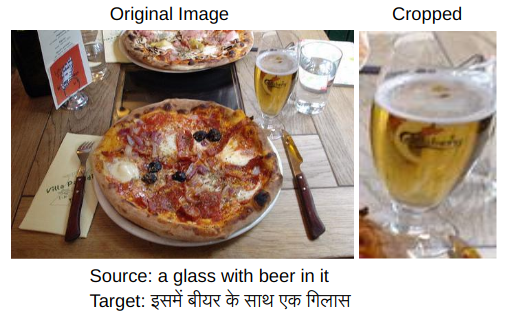}
\caption{An example of multimodal dataset}
\label{pic:example_img}
\end{figure}

Multimodal dataset consists of an image along with a description of certain rectangular portion of the image. We are given the coordinates of the portion. We aim to translate the description with help of the image.
An example of multimodal dataset is given in Figure \ref{pic:example_img}.
\begin{figure*}
\centering
\includegraphics[width=\textwidth]{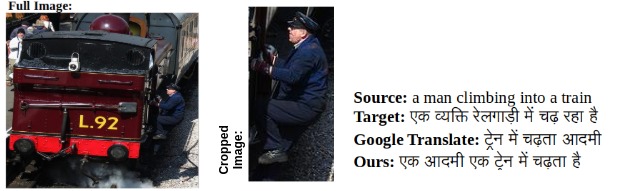}
\caption{An example of translation generated by the system. Here, the target is \textit{Ek vyakti railgari mein chad raha hai (A man climbing into train.)} The translation by Google NMT system is \textit{Train mein chadta Aadmi (Man climbs into train)}; whereas our NMT system translates it as: \textit{Ek aadmi ek train mein chadta hai (A man climbs into a train.)} }
\label{pic:example_result1}
\end{figure*} 

\begin{table*}[t]
\resizebox{\textwidth}{!}{%
\begin{tabular}{lllllll}
\hline
\multirow{2}{*}{\textbf{Team}} & \multicolumn{3}{c}{\textit{Evaluation} }                                                 & \multicolumn{3}{c}{\textit{Challenge}}                                                   \\ \cline{2-7} 
                      & \multicolumn{1}{c}{BLEU} & \multicolumn{1}{c}{RIBES} & \multicolumn{1}{c}{AMFM} & \multicolumn{1}{c}{BLEU} & \multicolumn{1}{c}{RIBES} & \multicolumn{1}{c}{AMFM} \\ \hline
Volta                 & 44.21                    & 0.818689                  & 0.835480                 & 52.02                    & 0.854139                  & 0.874220                 \\
iitp (Ours)           & 42.47                    & 0.807123                  & 0.819720                 & 37.50                    & 0.790809                  & 0.830230                 \\
CNLP-NITS             & 40.51                    & 0.803208                  & 0.820980                 & 39.28                    & 0.792097                  & 0.812360                 \\
CNLP-NITS             & 39.46                    & 0.802055                  & 0.823270                 & 33.57                    & 0.754141                  & 0.787320                 \\
Organizer             & 38.63                    & 0.767422                  & 0.772870                 & 20.34                    & 0.644230                  & 0.669760                 \\            
\end{tabular}% 
}
\caption{Details of obtained results by different submissions}
\label{tab:results}
\end{table*}

\subsection{Pre-processing} For text data, we lowercase all the utterances. Then, we jointly learn byte-pair-encoding \cite{sennrich-etal-2016-neural} combining both source and target with a vocabulary of 10,000. We treat the images by cropping a specified rectangular portions. This operation is used to discard the portions that do not contribute much to the translation performance. After we get those cropped-out images, we use the pre-trained VGG19-bn \cite{simonyan2015deep} to obtain the image representations. We use OpenNMT-py \cite{klein-etal-2017-opennmt} framework to perform this step.
\subsection{Training}

We use OpenNMT-py \cite{klein-etal-2017-opennmt} for our NMT systems. We use Bidirectional RNN encoder and doubly attentive RNN decoder \cite{calixto-etal-2017-doubly} for our experiments.  We train our system in two ways \textit{viz.} With pre-training, and Without pre-training.:

\begin{enumerate}
    \item \textbf{With pre-training}
    We pre-train one of our models on HindEnCorp dataset. This step does not use any visual features as the dataset used for pre-training is devoid of any visual features. After pre-training, we fine-tune the pre-trained model with VisualGenome dataset containing textual and visual features.
    \item \textbf{Without pre-training} We do not pre-train the model. We directly fine-tune the models on VisualGenome dataset which contains both text and associated image. Consequently, both textual and visual features are used.
\end{enumerate}

Following step is taken into account while doing inference step:\\
We take the best hypothesis from both the models and filter out any hypothesis containing \textless unk\textgreater  token. Then, we pick the hypothesis with best log-likelihood during generation.
\subsection{Hyper-parameters}
We set the word embedding size and size of RNN hidden states to 500. We set the batch size to 40 and train for a maximum 25 epochs. We restrict maximum source and target sequence length to 50. We use the Adam optimizer \cite{kingma2017adam} for optimization with $\beta_1=0.9$ and $\beta_2=0.999$. During training, we use 0.3 as dropout rate to avoid over-fitting. During generation of translation, we use 5 as the beam width. 

\section{Experimental Results}
We obtain impressive results on our submissions. There are two sets designed for evaluating our model, \textit{i) Evaluation set}, \textit{ii) Challenge set}. We evaluate our model on both of these test set and tabulate our results in Table \ref{tab:results}. We use different evaluation metrics (BLEU, RIBES, AMFM) to test our model. The results shown in the table are sorted according to the obtained BLEU scores.
As it can be seen from Table \ref{tab:results}, we obtain $42.47$ BLEU points and achieve second position in terms of BLEU on Evaluation set on multimodal task. Please refer to Figure \ref{pic:example_result1} for example of translation by our system.
We obtain 37.50 BLEU points on Challenge set. One reason for not so good results on Challenge set could be:
\begin{itemize}
    \item The challenge test set was created by searching for (particularly) ambiguous English words based on the embedding similarity and manually selecting those where the image helps to resolve the ambiguity. Hence, it is difficult to translate compared to the Evaluation set, which was randomly selected.
    \item Difference between utterance length during training and testing, i.e. while average length of Train, Evaluation and Validation set is 5 but average length of Challenge set is 6.
\end{itemize}
\section{Conclusion}
We participate in WAT-2021 Multimodal Translation Task for English to Hindi. We achieve good results on both the Challenge and Evaluation sets achieving $42.47$ and $37.50$ BLEU points, respectively. We rank second place on Evaluation set and third place on Challenge set on WAT-2021 Multimodal Translation Task for English to Hindi. In future, we would like to extend our work by training with additional monolingual data and better ways to incorporate multimodal features.

% \usepackage{multirow}
% \usepackage{graphicx}

%The acknowledgments should go immediately before the references. Do not number the acknowledgments section.

\bibliographystyle{acl_natbib}
\bibliography{anthology,acl2021}

\begin{thebibliography}{22}
\expandafter\ifx\csname natexlab\endcsname\relax\def\natexlab#1{#1}\fi

\bibitem[{Calixto and Liu(2017)}]{calixto-liu-2017-incorporating}
Iacer Calixto and Qun Liu. 2017.
\newblock \href {https://doi.org/10.18653/v1/D17-1105} {Incorporating global
  visual features into attention-based neural machine translation.}
\newblock In \emph{Proceedings of the 2017 Conference on Empirical Methods in
  Natural Language Processing}, pages 992--1003, Copenhagen, Denmark.
  Association for Computational Linguistics.

\bibitem[{Calixto et~al.(2017)Calixto, Liu, and
  Campbell}]{calixto-etal-2017-doubly}
Iacer Calixto, Qun Liu, and Nick Campbell. 2017.
\newblock \href {https://doi.org/10.18653/v1/P17-1175} {Doubly-attentive
  decoder for multi-modal neural machine translation}.
\newblock In \emph{Proceedings of the 55th Annual Meeting of the Association
  for Computational Linguistics (Volume 1: Long Papers)}, pages 1913--1924,
  Vancouver, Canada. Association for Computational Linguistics.

\bibitem[{Chen et~al.(2019)Chen, Li, Yu, Kholy, Ahmed, Gan, Cheng, and
  Liu}]{DBLP:journals/corr/abs-1909-11740}
Yen{-}Chun Chen, Linjie Li, Licheng Yu, Ahmed~El Kholy, Faisal Ahmed, Zhe Gan,
  Yu~Cheng, and Jingjing Liu. 2019.
\newblock \href {http://arxiv.org/abs/1909.11740} {{UNITER:} learning universal
  image-text representations}.
\newblock \emph{CoRR}, abs/1909.11740.

\bibitem[{Delbrouck and Dupont(2017)}]{delbrouck2017modulating}
Jean-Benoit Delbrouck and Stéphane Dupont. 2017.
\newblock \href {http://arxiv.org/abs/1712.03449} {Modulating and attending the
  source image during encoding improves multimodal translation}.

\bibitem[{Dutta~Chowdhury et~al.(2018)Dutta~Chowdhury, Hasanuzzaman, and
  Liu}]{dutta-chowdhury-etal-2018-multimodal}
Koel Dutta~Chowdhury, Mohammed Hasanuzzaman, and Qun Liu. 2018.
\newblock \href {https://doi.org/10.18653/v1/W18-3405} {Multimodal neural
  machine translation for low-resource language pairs using synthetic data}.
\newblock In \emph{Proceedings of the Workshop on Deep Learning Approaches for
  Low-Resource {NLP}}, pages 33--42, Melbourne. Association for Computational
  Linguistics.

\bibitem[{Gain et~al.(2021)Gain, Haque, and Ekbal}]{baban-gain-chat-ijcnn}
Baban Gain, Rejwanul Haque, and Asif Ekbal. 2021.
\newblock Not all contexts are important: The impact of effective context in
  conversational neural machine translation.
\newblock In \emph{2021 International Joint Conference on Neural Networks
  (IJCNN)}.

\bibitem[{Huang et~al.(2016)Huang, Liu, Shiang, Oh, and
  Dyer}]{huang-etal-2016-attention}
Po-Yao Huang, Frederick Liu, Sz-Rung Shiang, Jean Oh, and Chris Dyer. 2016.
\newblock \href {https://doi.org/10.18653/v1/W16-2360} {Attention-based
  multimodal neural machine translation}.
\newblock In \emph{Proceedings of the First Conference on Machine Translation:
  Volume 2, Shared Task Papers}, pages 639--645, Berlin, Germany. Association
  for Computational Linguistics.

\bibitem[{Kingma and Ba(2017)}]{kingma2017adam}
Diederik~P. Kingma and Jimmy Ba. 2017.
\newblock \href {http://arxiv.org/abs/1412.6980} {Adam: A method for stochastic
  optimization}.

\bibitem[{Klein et~al.(2017)Klein, Kim, Deng, Senellart, and
  Rush}]{klein-etal-2017-opennmt}
Guillaume Klein, Yoon Kim, Yuntian Deng, Jean Senellart, and Alexander Rush.
  2017.
\newblock \href {https://www.aclweb.org/anthology/P17-4012} {{O}pen{NMT}:
  Open-source toolkit for neural machine translation}.
\newblock In \emph{Proceedings of {ACL} 2017, System Demonstrations}, pages
  67--72, Vancouver, Canada. Association for Computational Linguistics.

\bibitem[{Laskar et~al.(2020)Laskar, Khilji, Pakray, and
  Bandyopadhyay}]{laskar-etal-2020-multimodal}
Sahinur~Rahman Laskar, Abdullah Faiz Ur~Rahman Khilji, Partha Pakray, and
  Sivaji Bandyopadhyay. 2020.
\newblock \href {https://www.aclweb.org/anthology/2020.wat-1.11} {Multimodal
  neural machine translation for {E}nglish to {H}indi}.
\newblock In \emph{Proceedings of the 7th Workshop on Asian Translation}, pages
  109--113, Suzhou, China. Association for Computational Linguistics.

\bibitem[{Li et~al.(2019)Li, Yatskar, Yin, Hsieh, and
  Chang}]{DBLP:journals/corr/abs-1908-03557}
Liunian~Harold Li, Mark Yatskar, Da~Yin, Cho{-}Jui Hsieh, and Kai{-}Wei Chang.
  2019.
\newblock \href {http://arxiv.org/abs/1908.03557} {Visualbert: {A} simple and
  performant baseline for vision and language}.
\newblock \emph{CoRR}, abs/1908.03557.

\bibitem[{Lin et~al.(2020)Lin, Meng, Su, Yin, Yang, Ge, Zhou, and
  Luo}]{Lin_2020}
Huan Lin, Fandong Meng, Jinsong Su, Yongjing Yin, Zhengyuan Yang, Yubin Ge, Jie
  Zhou, and Jiebo Luo. 2020.
\newblock \href {https://doi.org/10.1145/3394171.3413715} {Dynamic
  context-guided capsule network for multimodal machine translation}.
\newblock \emph{Proceedings of the 28th ACM International Conference on
  Multimedia}.

\bibitem[{Nakazawa et~al.(2021)Nakazawa, Nakayama, Ding, Dabre, Higashiyama,
  Mino, Goto, Pa, Kunchukuttan, Parida, Bojar, Chu, Eriguchi, Abe, and
  Oda}]{nakazawa-etal-2021-overview}
Toshiaki Nakazawa, Hideki Nakayama, Chenchen Ding, Raj Dabre, Shohei
  Higashiyama, Hideya Mino, Isao Goto, Win~Pa Pa, Anoop Kunchukuttan,
  Shantipriya Parida, Ondřej Bojar, Chenhui Chu, Akiko Eriguchi, Kaori Abe,
  and Sadao Oda, Yusuke~Kurohashi. 2021.
\newblock Overview of the 8th workshop on {A}sian translation.
\newblock In \emph{Proceedings of the 8th Workshop on Asian Translation},
  Bangkok, Thailand. Association for Computational Linguistics.

\bibitem[{Nakazawa et~al.(2020)Nakazawa, Nakayama, Ding, Dabre, Higashiyama,
  Mino, Goto, Pa~Pa, Kunchukuttan, Parida, Bojar, and
  Kurohashi}]{nakazawa-etal-2020-overview}
Toshiaki Nakazawa, Hideki Nakayama, Chenchen Ding, Raj Dabre, Shohei
  Higashiyama, Hideya Mino, Isao Goto, Win Pa~Pa, Anoop Kunchukuttan,
  Shantipriya Parida, Ond{\v{r}}ej Bojar, and Sadao Kurohashi. 2020.
\newblock \href {https://www.aclweb.org/anthology/2020.wat-1.1} {Overview of
  the 7th workshop on {A}sian translation}.
\newblock In \emph{Proceedings of the 7th Workshop on Asian Translation}, pages
  1--44, Suzhou, China. Association for Computational Linguistics.

\bibitem[{Parida et~al.(2019)Parida, Bojar, and
  Dash}]{hindi-visual-genome:2019}
Shantipriya Parida, Ond{\v{r}}ej Bojar, and Satya~Ranjan Dash. 2019.
\newblock {Hindi Visual Genome: A Dataset for Multimodal English-to-Hindi
  Machine Translation}.
\newblock \emph{Computaci{\'o}n y Sistemas}, 23(4):1499--1505.
\newblock Presented at CICLing 2019, La Rochelle, France.

\bibitem[{Sabour et~al.(2017)Sabour, Frosst, and Hinton}]{sabour2017dynamic}
Sara Sabour, Nicholas Frosst, and Geoffrey~E Hinton. 2017.
\newblock \href {http://arxiv.org/abs/1710.09829} {Dynamic routing between
  capsules}.

\bibitem[{Sanayai~Meetei et~al.(2019)Sanayai~Meetei, Singh, and
  Bandyopadhyay}]{sanayai-meetei-etal-2019-wat2019}
Loitongbam Sanayai~Meetei, Thoudam~Doren Singh, and Sivaji Bandyopadhyay. 2019.
\newblock \href {https://doi.org/10.18653/v1/D19-5224} {{WAT}2019:
  {E}nglish-{H}indi translation on {H}indi visual genome dataset}.
\newblock In \emph{Proceedings of the 6th Workshop on Asian Translation}, pages
  181--188, Hong Kong, China. Association for Computational Linguistics.

\bibitem[{Sennrich et~al.(2016)Sennrich, Haddow, and
  Birch}]{sennrich-etal-2016-neural}
Rico Sennrich, Barry Haddow, and Alexandra Birch. 2016.
\newblock \href {https://doi.org/10.18653/v1/P16-1162} {Neural machine
  translation of rare words with subword units}.
\newblock In \emph{Proceedings of the 54th Annual Meeting of the Association
  for Computational Linguistics (Volume 1: Long Papers)}, pages 1715--1725,
  Berlin, Germany. Association for Computational Linguistics.

\bibitem[{Simonyan and Zisserman(2015)}]{simonyan2015deep}
Karen Simonyan and Andrew Zisserman. 2015.
\newblock \href {http://arxiv.org/abs/1409.1556} {Very deep convolutional
  networks for large-scale image recognition}.

\bibitem[{Su et~al.(2021)Su, Chen, Jiang, Zhou, Lin, Ge, Wu, and
  Lai}]{SU202147}
Jinsong Su, Jinchang Chen, Hui Jiang, Chulun Zhou, Huan Lin, Yubin Ge,
  Qingqiang Wu, and Yongxuan Lai. 2021.
\newblock \href {https://doi.org/https://doi.org/10.1016/j.ins.2020.11.024}
  {Multi-modal neural machine translation with deep semantic interactions}.
\newblock \emph{Information Sciences}, 554:47--60.

\bibitem[{Su et~al.(2018)Su, Fan, Bach, Kuo, and
  Huang}]{DBLP:journals/corr/abs-1811-11365}
Yuanhang Su, Kai Fan, Nguyen Bach, C.{-}C.~Jay Kuo, and Fei Huang. 2018.
\newblock \href {http://arxiv.org/abs/1811.11365} {Unsupervised multi-modal
  neural machine translation}.
\newblock \emph{CoRR}, abs/1811.11365.

\bibitem[{Wang et~al.(2017)Wang, Tu, Way, and
  Liu}]{wang-etal-2017-exploiting-cross}
Longyue Wang, Zhaopeng Tu, Andy Way, and Qun Liu. 2017.
\newblock \href {https://doi.org/10.18653/v1/D17-1301} {Exploiting
  cross-sentence context for neural machine translation}.
\newblock In \emph{Proceedings of the 2017 Conference on Empirical Methods in
  Natural Language Processing}, pages 2826--2831, Copenhagen, Denmark.
  Association for Computational Linguistics.

\end{thebibliography}

%\appendix

\end{document}